\pdfoutput=1

\documentclass[11pt]{article}

\usepackage{acl}

\usepackage{times}
\usepackage{latexsym}

\usepackage[T1]{fontenc}

\usepackage[utf8]{inputenc}

\usepackage{microtype}

\usepackage{inconsolata}

\usepackage{rotating}
\usepackage{blindtext}
\usepackage{amsmath}
\usepackage{amsfonts}
\usepackage{graphicx}
\usepackage{mathtools}
\usepackage{booktabs}
\usepackage{relsize}
\usepackage{float}
\usepackage{graphics}
\usepackage{multirow}
\usepackage{algorithm}
\usepackage{algpseudocode}
\usepackage[utf8]{inputenc}
\usepackage{siunitx}
\newrobustcmd\B{\DeclareFontSeriesDefault[rm]{bf}{b}\bfseries}

\DeclareUnicodeCharacter{01B0}{\`u}
\DeclareUnicodeCharacter{01A1}{\`o}

\usepackage{threeparttable}
\usepackage{booktabs, caption, makecell}

\usepackage[title]{appendix}
\usepackage[cal=cm]{mathalfa}
\usepackage[colorinlistoftodos]{todonotes}
\usepackage{multirow}
\usepackage[export]{adjustbox}
\usepackage{subcaption}
\usepackage{pifont}
\newcommand{\ignore}[1]{}

\usepackage{microtype}
\usepackage[subtle]{savetrees}
\title{Investigating Multi-Pivot Ensembling \\with Massively Multilingual Machine Translation Models}

\author{Alireza Mohammadshahi~\thanks{~Work done while working at University of Zurich. Currently co-founder of Leeroo.}~~$^{1,2}$ ~~Jannis Vamvas ~$^{1}$ ~~Rico Sennrich~$^{1}$
\vspace{0.1cm}\\ 
 $^1$ University of Zurich ~~~~~ $^2$ EPFL\\
 \texttt{alireza.mohammadshahi@epfl.ch}\\\texttt{\{vamvas,sennrich\}@cl.uzh.ch}
}

\begin{document}
\maketitle

\begin{abstract}

Massively multilingual machine translation models allow for the translation of a large number of languages with a single model, but have limited performance on low- and very-low-resource translation directions.
Pivoting via high-resource languages remains a strong strategy for low-resource directions, and in this paper we revisit ways of pivoting through multiple languages.
Previous work has used a simple averaging of probability distributions from multiple paths, but we find that this performs worse than using a single pivot, and exacerbates the hallucination problem because the same hallucinations can be probable across different paths. We also propose MaxEns, a novel combination strategy that makes the output biased towards the most confident predictions, hypothesising that confident predictions are less prone to be hallucinations.
We evaluate different strategies on the FLORES benchmark for 20 low-resource language directions, demonstrating that MaxEns improves translation quality for low-resource languages while reducing hallucination in translations, compared to both direct translation and an averaging approach. On average, multi-pivot strategies still lag behind using English as a single pivot language, raising the question of how to identify the best pivoting strategy for a given translation direction.\footnote{The implementation is publicly available at \url{https://github.com/ZurichNLP/MultiPivotNMT}.}

\end{abstract}

\section{Introduction}

Early work on multilingual neural machine translation~(NMT) has explored combining source segments in different source languages~\cite{zoph-knight-2016-multi,firat-etal-2016-multi}, an idea that is also compatible with pivoting through intermediate languages.
For example, one could translate from Dutch to Ukrainian by first translating the Dutch source to English and Russian, and then making a combined prediction to Ukrainian.
In the simplest case, this combination is achieved by predicting probability distributions for each source language and averaging these predictions in an ensemble-like manner \cite{firat-etal-2016-multi}.

With massively multilingual NMT models~\cite{team2022language,mohammadshahi-etal-2022-small,goyal-etal-2022-flores,wenzek-etal-2021-findings,zhang-etal-2020-improving,fan2020englishcentric,aharoni-etal-2019-massively,arivazhagan2019massively}, one can in principle translate directly in any translation direction. While early models relied on zero-shot generalization for many directions, recent improvements include massive data collection efforts~\cite{schwenk-etal-2021-ccmatrix,el-kishky-etal-2020-ccaligned} and synthetic data creation via back-translation~\cite{edunov-etal-2018-understanding,sennrich-etal-2016-improving}.
However, these models still have low performance on many low-resource translation directions\footnote{SentencePiece BLEU of 63\% translation directions in M2M-100 is lower than 12 \citep{mohammadshahi-etal-2022-compressed}. } and pivot-translation via high-resource languages remains a strong baseline.
\citet{fan2020englishcentric} also investigate the combination of multiple translation paths, which they call \textit{multi-source self-ensemble}, that slightly improves over the direct translation and a single pivot for zero-shot language pairs.

In this paper, we investigate this multi-source self-ensembling strategy more closely, with a focus on preventing completely defunct translations such as hallucinations.
However, we find that simple averaging is sub-optimal and may increase the number of hallucinations in the output, a typical failure case in low-resource settings.
We relate this to a recent finding that hallucinations are \textit{sticky}, meaning that different models trained on the same data and architecture may produce similar hallucinations \cite{guerreiro2023hallucinations}.
We also find evidence of such stickiness when combining multiple translation paths, and propose a new ensembling strategy that, instead of averaging probabilities, picks the output with the maximum probability across different paths: \textbf{MaxEns}.
This is partially inspired by the finding that model confidence is a good heuristic for avoiding hallucinations, which tend to be low-confidence predictions \cite{guerreiro-etal-2023-looking}.

We perform experiments on the FLORES benchmark~\cite{goyal-etal-2022-flores} for 20 low-resource translation directions by using two massively multilingual NMT models, SMaLL100~\cite{mohammadshahi-etal-2022-small} and M2M100~\cite{fan2020englishcentric}. 
Our results show that while the average ensemble outperforms the direct translation, it still underperforms using only English as a pivot, both in terms of spBLEU and the number of hallucinations.
MaxEns performs significantly better than the averaging strategy for both translation performance and hallucination. Specially, MaxEns has competitive translation performance with English pivoting on average, but still lags behind it on the hallucination performance.
To sum up, our contributions are:
\begin{itemize}
    \item We explore why a naive multi-pivot strategy with massively multilingual models can underperform single-pivot translation. Then, we propose MaxEns, a more robust ensembling technique for multi-pivot translation with multilingual NMT models.
    \item We evaluate different ensembling strategies on 20 low-resource translation directions of FLORES benchmark, and demonstrate that multi-pivot ensembling still lags behind the English pivoting.
\end{itemize}

\section{Related Work}
\label{relatedwork}

Several approaches exploited different multi-pivoting methods to improve the performance of NMT models, specifically for low-resource language directions~\cite{machacek-etal-2023-robustness,dabre2021simultaneous,kim-etal-2019-pivot,ijcai2017p555,firat-etal-2016-zero}. \citet{machacek-etal-2023-robustness} analyzed the robustness of multi-source NMT in transcription errors. \citet{dabre2021simultaneous} improved the performance of simultaneous NMT by translating the source language into pivot languages, then applying the multi-source translation method~\cite{zoph-knight-2016-multi}.~\citet{firat-etal-2016-zero} proposed a novel zero-resource translation approach by exploiting the multi-way multilingual NMT model, introduced by~\citet{firat-etal-2016-multi}, and improved the performance over traditional pivot-based translation~\cite{wu-wang-2007-pivot,utiyama-isahara-2007-comparison}.~\citet{ijcai2017p555} introduced the pivot-based NMT model by jointly training source-to-pivot and pivot-to-target directions.~\citet{currey-heafield-2019-zero} proposed an alternative method by applying a monolingual pivot-language data for zero-resource NMT via back-translation \cite{sennrich-etal-2016-improving}.

\section{Ensembling Methods}
When performing direct translation, the score of a translation $Y$ given a source sequence $X_{src}$ is computed as follows:
\begin{equation}
s(Y;X_{src}) = \sum_{i=1}^{|Y|} \log~p(y_i|y_{<i},X_{src})
\end{equation}
where $p(y_i|y_{<i},X_{src})$ is the predicted probability of the $i$-th target token $y_i$ given the previous tokens $y_{<i}$ and the source sequence $X_{src}$.

For multi-pivot ensembling, we select a set of pivot languages~$M=\{\mu_1,\mu_2,...,\mu_K\}$ and generate the corresponding pivot translations $X_M=\{X_{\mu_1},X_{\mu_2},...,X_{\mu_K}\}$.
The final translation is generated by ensembling predictions, conditioned on the individual pivot translations.

 In the following, we describe two approaches for such an ensembling: the multilingual averaging method and our MaxEns approach.

\paragraph{Multilingual Average~(MultiAvg).} Inspired by \citet{fan2020englishcentric,firat-etal-2016-zero}, we average the predicted probabilities of a token $y_i$ across all pivot languages:\footnote{We tried both averaging probabilities and log-probabilities in preliminary experiments, and averaging probabilities worked better in terms of translation performance and hallucination.}

\begin{equation}
    s(Y;X_M) = \sum_{i=1}^{|Y|} \log \frac{1}{|M|}\sum_{k=1}^{|M|} ~p(y_i|y_{<i},X_{\mu_k}).
\end{equation}
where $|Y|$ and $|M|$ are the number of target tokens and pivots, respectively. 

\begin{figure*}[t!]
\centering
\includegraphics[width=\textwidth]{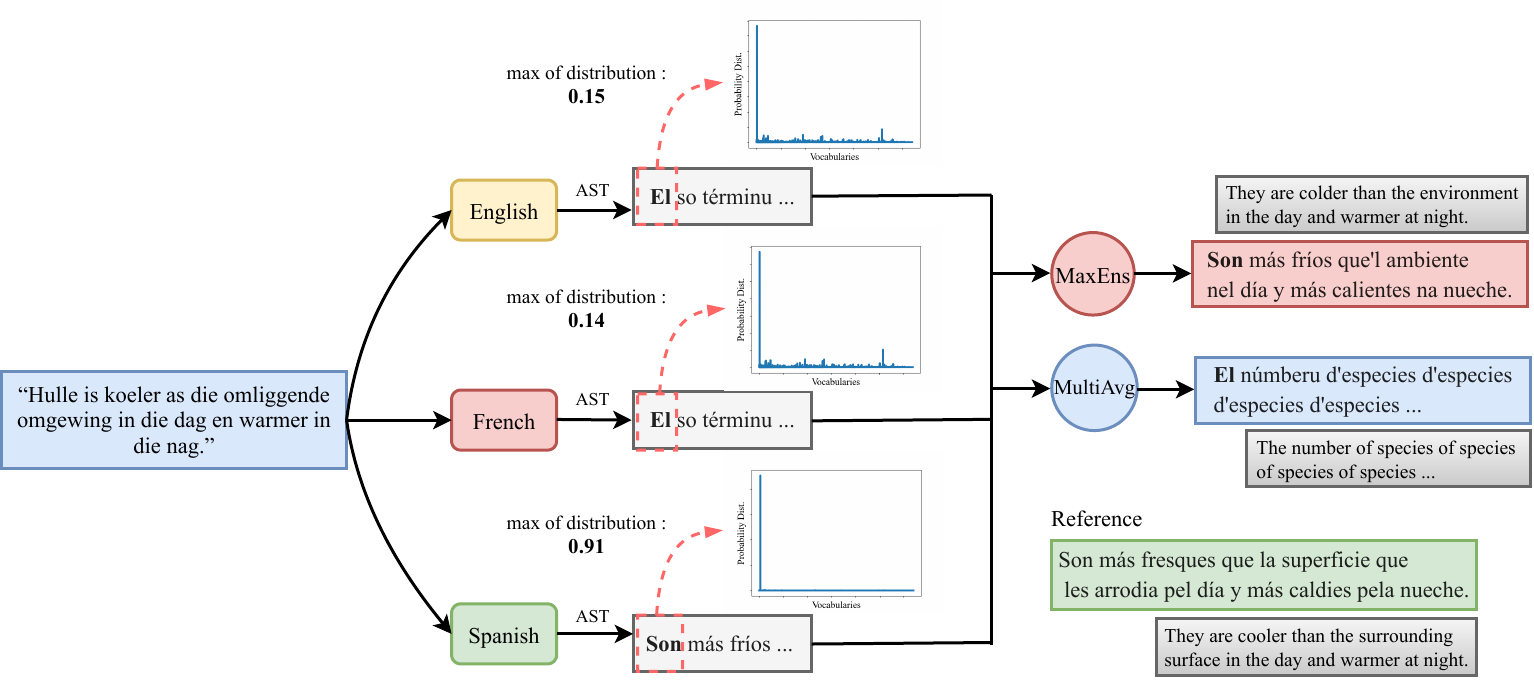}
\caption{
A sample translation of Afrikaans to Asturian by using SMaLL100. Translations of individual pivots are shown in the middle, output translations of MaxEns and MultiAvg on the right. MaxEns method eliminates the hallucination, as it follows the more confident pivot~(here, Spanish). Glosses in English are presented within gray boxes.
}
\label{fig:sample}
\end{figure*}

\paragraph{Maximum Ensemble~(MaxEns).} As novel combination strategy that biases the prediction towards the more confident pivot, we propose the following approach:

\begin{equation}
    s(Y;X_M) = \sum_{i=1}^{|Y|} \max_{k=1}^{|M|}[\log ~p(y_i|y_{<i},X_{\mu_k})].
\end{equation}
where it chooses the maximum score between predictions of pivots for token $y_i$. Intuitively, MaxEns selects the most confident pivot language when generating token $y_i$.

\section{Results and Discussion}
\label{sec:results}

\subsection{Experimental Setup}
\label{sec:result:1}
\paragraph{Models.} We used M2M100 and SMaLL100 as our massively multilingual NMT models. M2M100 is trained on large-scale multilingual corpora~\cite{schwenk-etal-2021-ccmatrix,el-kishky-etal-2020-ccaligned} with a novel data mining procedure, that uses language similarities. We exploit M2M100 variant with 418M parameters. SMaLL100~\cite{mohammadshahi-etal-2022-small} is a distilled version of M2M100 12B with 330M parameters. It has been trained with uniform sampling across all language pairs on nearly 6\% of M2M100 pre-training dataset, and achieved competitive performance with M2M100 with 1.2B parameters.

\paragraph{Evaluation Setting.} Inspired by \citet{fan2020englishcentric}, we use the FLORES-101 benchmark~\cite{goyal-etal-2022-flores}. It contains 3,001 sentences derived from English Wikipedia, and translated into 101 languages by human.
We use the \textit{devtest} subset for the evaluation. To better understand the effect of multilingual pivoting, we chose five low-resource (or very low) languages from different branches of Indo-European, including Germanic, Romance, Slavic, Indo-Aryan, and Iranian. These languages are Afrikaans, Asturian, Croatian, Urdu, and Pashto. We evaluate on all permutations of these languages, which results in 20 language pairs. 
As pivot languages, we use English, Spanish and French.
English has the largest amount of bitext overall in the training data of M2M100, and Spanish and French have the largest amount of bitext with English~\cite{fan2020englishcentric}.

\paragraph{Evaluation Metrics.} spBLEU\footnote{BLEU is computed after tokenization with SentencePiece with 256K tokens \cite{goyal-etal-2022-flores}.} is used to measure the translation performance~\cite{goyal-etal-2022-flores}. For the hallucination measurement, we apply a coarse estimation method inspired by \citet{lee2019hallucinations,muller-sennrich-2021-understanding}, counting the proportion of sentences with ChrF~\cite{popovic-2015-chrf}\footnote{sacrebleu 2.3.1~\cite{post-2018-call} with ChrF3 is used.} less than 20.\footnote{Threshold based on manual inspection.}  Additionally, we use top n-gram (TNG)~\cite{guerreiro-etal-2023-looking,raunak2022salted,raunak-etal-2021-curious} for detecting oscillatory hallucinations.\footnote{We follow \citet{guerreiro-etal-2023-looking} and use $n=4$ and $t=2$.} We apply significance testing with $p=0.05$.\footnote{Paired bootstrap resampling~\cite{koehn-2004-statistical} with sacrebleu.} Beam size 5 is used for inference.

\begin{table}[t]
	\centering
	\tabcolsep=0.1cm
	\begin{adjustbox}{width=\linewidth}
	\begin{tabular}{lS[table-format=2.1]S[table-format=2.1]S[table-format=2.1]S[table-format=2.1]}
    \toprule
    Language Pairs & \multicolumn{1}{c}{Direct} & \multicolumn{1}{c}{MultiAvg} & \multicolumn{1}{c}{MaxEns} & \multicolumn{1}{c}{EN Pivot}\\
    \midrule
Afrikaans-Asturian & 20.6 &21.8 & \B 22.5 & 21.0 \\
Afrikaans-Croatian & 22.1 &22.5 & 22.4 & \B 22.8 \\
Afrikaans-Urdu &14.0 & 13.8 & 13.9 & \B 14.4 \\
Afrikaans-Pashto & \B 5.9 & \B 5.8 & \B 6.0 & \B 6.0 \\
Asturian-Afrikaans & 18.5 & 19.9 & 19.8 & \B 21.0 \\
Asturian-Croatian & 16.1 & \B 20.4 & 20.0 & \B 20.4 \\
Asturian-Urdu & 8.4 & 11.8 & 11.8 & \B 12.6 \\
Asturian-Pashto & 3.6 & 5.0 & 5.0 & \B 5.4 \\
Croatian-Afrikaans & 20.5 & 20.7 & 20.8 & \B 21.2 \\
Croatian-Asturian & 19.7 & 20.8 & \B 21.6 & 19.7 \\
Croatian-Urdu & \B 13.5 & 13.1 & 12.9 & 13.2 \\
Croatian-Pashto & 5.0 & 5.2 & 5.2 & \B 5.6 \\
Urdu-Afrikaans & 12.1 & 13.0 & 13.2 & \B 13.8 \\
Urdu-Asturian & 7.7 & 11.7 & \B 12.8 & 12.2 \\
Urdu-Croatian & 11.2 & \B 12.0 & \B 11.9 & \B 12.2 \\
Urdu-Pashto & \B 4.7 & \B 4.4 & 4.2 & \B 4.6 \\
Pashto-Afrikaans & 10.2 & 11.0 & 11.0 & \B 11.4 \\
Pashto-Asturian & 6.9 & 9.9 & \B 10.9 & 10.4 \\
Pashto-Croatian & 8.7 & \B 9.9 & \B 10.0 & \B 9.8 \\
Pashto-Urdu & \B 10.0 & 9.2 & 9.2 & 9.5 \\
\addlinespace
Average & 12.0 & 13.1 & \B 13.3 & \B 13.4 \\
		\bottomrule
	    \end{tabular}
	\end{adjustbox}
	\caption{\label{tab:spbleu} 
	   Average spBLEU (higher is better) of different pivoting methods for M2M100 and SMaLL100 on selected language pairs of FLORES-101. Best systems (not significantly outperformed by any other) in bold.
        }
\end{table}

\begin{table}[t]
	\centering
	\tabcolsep=0.1cm
	\begin{adjustbox}{width=\linewidth}
	\begin{tabular}{lS[table-format=2.1]S[table-format=2.1]S[table-format=2.1]S[table-format=2.1]}
    \toprule
    Language Pairs & \multicolumn{1}{c}{Direct} & \multicolumn{1}{c}{MultiAvg} & \multicolumn{1}{c}{MaxEns} & \multicolumn{1}{c}{EN Pivot}\\
    \midrule
Afrikaans-Asturian &5.9 &6.4 &4.0 &4.3 \\
Afrikaans-Croatian &1.7 &2.2 &2.2 &1.9 \\
Afrikaans-Urdu &11.3 &13.9 &13.4 &11.1 \\
Afrikaans-Pashto &49.5 &54.1 &53.2 &49.8 \\
Asturian-Afrikaans &8.8 &6.2 &7.8 &2.1 \\
Asturian-Croatian &19.5 &5.3 &8.2 &3.5 \\
Asturian-Urdu &40.7 &23.4 &22.9 &18.0 \\
Asturian-Pashto &66.7 &62.2 &62.2 &55.9 \\
Croatian-Afrikaans &1.1 &1.5 &1.2 &1.3 \\
Croatian-Asturian &6.1 &6.3 &3.8 &5.0 \\
Croatian-Urdu &13.2 &16.5 &16.3 &12.7 \\
Croatian-Pashto &54.7 &58.4 &56.6 &53.1 \\
Urdu-Afrikaans &9.8 &9.2 &9.4 &4.9 \\
Urdu-Asturian &31.5 &19.9 &14.7 &12.5 \\
Urdu-Croatian &14.1 &15.4 &15.5 &11.8 \\
Urdu-Pashto &56.8 &66.0 &66.8 &60.1 \\
Pashto-Afrikaans &9.1 &10.5 &10.0 &6.6 \\
Pashto-Asturian &26.9 &21.2 &15.7 &15.6 \\
Pashto-Croatian &18.4 &18.5 &18.7 &16.7 \\
Pashto-Urdu &24.7 &33.2 &33.3 &28.9 \\
\addlinespace
Average &23.5 &22.5 &21.8 &18.8 \\
		\bottomrule
	    \end{tabular}
	\end{adjustbox}
	\caption{\label{tab:chrf} 
	  Average percentage~(100\%) of hallucinations (chrF < 20; lower is better) of different pivoting methods for M2M100 and SMaLL100 on selected language pairs of FLORES-101.
        }
\end{table}

\subsection{Results \& Discussion} 

Figure~\ref{fig:sample} illustrates an example of multi-pivot translation of Afrikaans to Asturian by using English, French, and Spanish as pivots. Translations via English and French pivots are hallucinations, while the translation via the Spanish pivot is more related to the reference translation. The output of MaxEns approach is closer to the translation achieved by using Spanish as the pivot language, since the NMT model is more confident for this pivot~(maximum of output probability distributions for the first token of English, French, and Spanish pivots are 0.15, 0.14, and 0.91, respectively). In contrast, the output of the MultiAvg method is a hallucination. \\
Tables~\ref{tab:spbleu} and \ref{tab:chrf} show translation and hallucination performances on 20 language directions, respectively.\footnote{Average scores of M2M100 and SMaLL100 are shown in Table~\ref{tab:spbleu} and \ref{tab:chrf}. Individual scores are provided in Appendix~\ref{app:scores}.} TNG scores for measuring oscillatory hallucinations are provided in Appendix~\ref{app:tng}. MultiAvg approach achieves better translation performance and lower hallucination compared to the direct translation.
However, MultiAvg method lags behind the English pivoting approach in terms of both translation quality~(13.4 vs.\ 13.1) and the occurrence of hallucinations~(18.8\% vs.\ 22.5\%).\footnote{4.5\% vs. 7.3\% based on the TNG metric, as shown in Table~\ref{app:tab:chrf-tng-avg}.} \\
Applying the MaxEns method instead for combining pivots tightens this gap, and leads to better translation and reduces the occurrence of hallucinations. Specifically, MaxEns reaches competitive translation quality with English pivoting on average (13.3 vs. 13.4), while still underperforming on hallucinations, as most of the parallel sentences of pre-training data for M2M100 and SMaLL100 are paired with English. 

In general, the optimal strategy differs across translation directions, highlighting the potential for future research on determining the most effective translation strategy for each direction without depending on the development data for each.
\section{Conclusion}

We investigate more closely the multi-source self-ensembling method of \citet{fan2020englishcentric} for combining multiple translation paths to improve translations of low-resource (or very-low) language pairs. Specifically, this approach~(named MultiAvg, here) averages the predictions of probability distributions of each source language in an ensemble-like manner. We evaluated it on 20 low-resource language pairs of FLORES-101 benchmark by using two massively multilingual NMT models, SMALL100 and M2M100. The MultiAvg method performs better than direct translation in terms of both translation quality and hallucinations, however it lags behind applying only English as pivot. Then, we proposed MaxEns method, a novel combination method that chooses the maximum of prediction probabilities of pivots for each designated target token. This approach results in a better translation quality compared to MultiAvg, while reducing hallucinations. On average, it achieves competitive performance with English pivoting with regard to the translation quality metric, but performs worse with regard to the hallucination metric. The most effective translation strategy varies depending on the translation direction, suggesting the need for future research to identify the optimal strategy for each direction independently of the specific development data. We hope our findings are a starting point for the broader integration of ensemble techniques within the context of massively multilingual NMT.

The insights of our experiments, specifically the stickiness of hallucinations with different inputs, have inspired our follow-up work on source-contrastive decoding \cite{sennrich-etal-2024-mitigating}, which has empirically shown to be an effective strategy to mitigate hallucinations. Future work could revisit multi-pivot ensembling in combination with source-contrastive decoding.
\section*{Limitations}

We apply our method to two common massively multilingual NMT models, SMALL100 and M2M100; future work can extend our work to more recent large models e.g. NLLB200~\cite{team2022language} and LLMs~\cite{touvron2023llama,workshop2022bloom}. We tested our approach on a subset of 20 low-resource language directions, future research can study the method for further language directions, including medium-resource language pairs.
\section*{Acknowledgement}

This work was funded by the Swiss National Science Foundation (project MUTAMUR;  no.~176727/213976).

\bibliographystyle{acl_natbib}
\bibliography{acl2023,anthology}

\renewcommand\thesection{\Alph{section}}
\renewcommand\thesubsection{\thesection.\Alph{subsection}}
\setcounter{section}{0}
\onecolumn

\begin{appendices}

\section{Individual Scores of M2M100 and SMaLL100 Models}
\label{app:scores}

\subsection{M2M100 Results}
~~
\begin{table}[!htb]
	\centering
	\tabcolsep=0.1cm
	\begin{adjustbox}{width=0.48\linewidth}
	\begin{tabular}{lS[table-format=2.1]S[table-format=2.1]S[table-format=2.1]S[table-format=2.1]}
    \toprule
    Language Pairs & \multicolumn{1}{c}{Direct} & \multicolumn{1}{c}{MultiAvg} & \multicolumn{1}{c}{MaxEns} & \multicolumn{1}{c}{EN Pivot}\\
    \midrule
Afrikaans-Asturian &19.3 &20.2 & \B 21.0 &20.2 \\
Afrikaans-Croatian & \B 20.8 & \B 21.1 & \B 21.0 & \B 21.4 \\
Afrikaans-Urdu &14.0 &13.6 &13.8 & \B 14.4 \\
Afrikaans-Pashto & \B 5.4 & \B 5.4 & \B 5.5 & \B 5.6 \\
Asturian-Afrikaans &14.2 &16.5 &16.1 & \B 18.0 \\
Asturian-Croatian &11.1 & \B 19.0 &18.3 & \B 19.4 \\
Asturian-Urdu &6.3 & 11.4 & 11.4 & \B 12.6 \\
Asturian-Pashto &2.4 & 4.4 & 4.5 & \B 5.0 \\
Croatian-Afrikaans &17.6 & \B 17.9 & \B 17.9 & \B 18.2 \\
Croatian-Asturian & 18.8 & 19.5 & \B 20.6 & 19.1 \\
Croatian-Urdu & \B 13.6 & 13.3 & 13.1 & \B 13.4 \\
Croatian-Pashto &4.4 &4.9 &4.9 & \B 5.4 \\
Urdu-Afrikaans &9.0 &9.8 &10.0 & \B 10.6 \\
Urdu-Asturian &7.1 &9.9 & \B 10.8 & \B 10.9 \\
Urdu-Croatian &8.9 & \B 10.0 & \B 9.8 & \B 10.1 \\
Urdu-Pashto & \B 4.2 & 3.8 & 3.6 & \B 4.2 \\
Pashto-Afrikaans &8.3 & \B 9.3 & 8.9 & \B 9.3 \\
Pashto-Asturian &7.8 &9.6 & \B 10.4 & 9.7 \\
Pashto-Croatian & 8.0 & \B 9.0 & \B 9.0 & 8.5 \\
Pashto-Urdu & \B 9.8 & 8.9 & 8.9 & 9.0 \\
\addlinespace
Average &10.6 &11.9 &12.0 & \B 12.2 \\
		\bottomrule
	    \end{tabular}
	\end{adjustbox}
	\caption{\label{app:tab:spbleu} 
	   spBLEU (higher is better) of different pivoting methods for M2M100 model on selected language pairs of FLORES-101~\cite{goyal-etal-2022-flores} benchmark. Best systems (not significantly outperformed by any other) in bold.
        }
\end{table}
~~
\begin{table}[!htb]
	\centering
	\tabcolsep=0.1cm
	\begin{adjustbox}{width=0.48\linewidth}
	\begin{tabular}{lS[table-format=2.1]S[table-format=2.1]S[table-format=2.1]S[table-format=2.1]}
    \toprule
    Language Pairs & \multicolumn{1}{c}{Direct} & \multicolumn{1}{c}{MultiAvg} & \multicolumn{1}{c}{MaxEns} & \multicolumn{1}{c}{EN Pivot}\\
    \midrule
Afrikaans-Asturian &7.0 &8.8 &5.0 &3.6 \\
Afrikaans-Croatian &2.6 &2.6 &2.4 &2.3 \\
Afrikaans-Urdu &12.2 &15.5 &14.5 &11.5 \\
Afrikaans-Pashto &53.6 &59.7 &57.0 &54.8 \\
Asturian-Afrikaans &15.7 &10.0 &13.0 &2.8 \\
Asturian-Croatian &35.0 &7.3 &12.6 &4.4 \\
Asturian-Urdu &55.8 &26.7 &27.3 &18.4 \\
Asturian-Pashto &73.4 &68.0 &67.8 &58.6 \\
Croatian-Afrikaans &1.4 &1.9 &1.5 &1.7 \\
Croatian-Asturian &6.5 &8.4 &4.5 &3.9 \\
Croatian-Urdu &13.3 &17.0 &16.4 &13.6 \\
Croatian-Pashto &59.1 &62.0 &58.8 &57.5 \\
Urdu-Afrikaans &15.2 &14.5 &14.8 &7.7 \\
Urdu-Asturian &38.9 &28.0 &21.5 &15.9 \\
Urdu-Croatian &20.3 &22.3 &22.2 &17.9 \\
Urdu-Pashto &60.2 &72.7 &73.3 &65.1 \\
Pashto-Afrikaans &11.5 &12.8 &12.7 &9.0 \\
Pashto-Asturian &23.1 &25.1 &17.9 &18.0 \\
Pashto-Croatian &20.5 &21.6 &22.5 &21.1 \\
Pashto-Urdu &26.6 &35.9 &35.5 &31.9 \\
\addlinespace 
Average &27.6 &26.0 &25.0 &21.0 \\
\bottomrule
	    \end{tabular}
	\end{adjustbox}
	\caption{\label{app:tab:chrf} 
	   The percentage of hallucinations (chrF < 20; lower is better) of different pivoting methods for M2M100 model on selected language pairs of FLORES-101~\cite{goyal-etal-2022-flores} benchmark.
        }
\end{table}

\subsection{SMaLL100 Results}
~~

\begin{table}[!htb]
	\centering
	\tabcolsep=0.1cm
	\begin{adjustbox}{width=0.48\linewidth}
	\begin{tabular}{lS[table-format=2.1]S[table-format=2.1]S[table-format=2.1]S[table-format=2.1]}
    \toprule
    Language Pairs & \multicolumn{1}{c}{Direct} & \multicolumn{1}{c}{MultiAvg} & \multicolumn{1}{c}{MaxEns} & \multicolumn{1}{c}{EN Pivot}\\
    \midrule
Afrikaans-Asturian & 22.0 & 23.4 & \B 24.0 & 21.7 \\
Afrikaans-Croatian &23.5 & \B 23.9 & \B 23.8 & \B 24.1 \\
Afrikaans-Urdu &13.9 &14.0 &14.0 & \B 14.4 \\
Afrikaans-Pashto & \B 6.4 & 6.1 & \B 6.4 & \B 6.4 \\
Asturian-Afrikaans &22.8 & \B 23.3 & \B 23.4 & \B 23.8 \\
Asturian-Croatian &21.1 & \B 21.8 & \B 21.6 & \B 21.4 \\
Asturian-Urdu &10.5 &12.1 &12.2 & \B 12.5 \\
Asturian-Pashto &4.8 & \B 5.6 & \B 5.5 & \B 5.7 \\
Croatian-Afrikaans &23.4 &23.4 & 23.7 & \B 24.2 \\
Croatian-Asturian &20.6 & \B 22.1 & \B 22.5 &20.2 \\
Croatian-Urdu & \B 13.3 &12.8 &12.6 & \B 13.0 \\
Croatian-Pashto & \B 5.6 & \B 5.4 & \B 5.4 & \B 5.8 \\
Urdu-Afrikaans & 15.1 & 16.1 & 16.3 & \B 17.0 \\
Urdu-Asturian & 8.3 & 13.4 & \B 14.8 & 13.4 \\
Urdu-Croatian &13.4 &14.0 &13.9 &14.3 \\
Urdu-Pashto & \B 5.1 & \B 5.0 & \B 4.8 & \B 5.0 \\
Pashto-Afrikaans & 12.0 & 12.7 & 12.9 & \B 13.5 \\
Pashto-Asturian & 6.0 & 10.2 & \B 11.3 & \B 11.1 \\
Pashto-Croatian &9.4 & \B 10.8 & \B 11.0 & \B 11.0 \\
Pashto-Urdu & \B 10.2 &9.5 &9.4 &9.9 \\
\addlinespace
Average &13.4 &14.3 & \B 14.5 & \B 14.4 \\
		\bottomrule
	    \end{tabular}
	\end{adjustbox}
	\caption{\label{app:tab:spbleu2} 
	   spBLEU (higher is better) of different pivoting methods for SMaLL100 model on selected language pairs of FLORES-101~\cite{goyal-etal-2022-flores} benchmark. Best systems (not significantly outperformed by any other) in bold.
        }
\end{table}

~~

\begin{table}[!htb]
	\centering
	\tabcolsep=0.1cm
	\begin{adjustbox}{width=0.48\linewidth}
	\begin{tabular}{lS[table-format=2.1]S[table-format=2.1]S[table-format=2.1]S[table-format=2.1]}
    \toprule
    Language Pairs & \multicolumn{1}{c}{Direct} & \multicolumn{1}{c}{MultiAvg} & \multicolumn{1}{c}{MaxEns} & \multicolumn{1}{c}{EN Pivot}\\
    \midrule
Afrikaans-Asturian &4.7 &4.1 &2.9 &5.0 \\
Afrikaans-Croatian &0.9 &1.9 &2.1 &1.5 \\
Afrikaans-Urdu &10.4 &12.3 &12.2 &10.8 \\
Afrikaans-Pashto &45.4 &48.4 &49.4 &44.8 \\
Asturian-Afrikaans &1.9 &2.5 &2.6 &1.5 \\
Asturian-Croatian &4.1 &3.3 &3.9 &2.7 \\
Asturian-Urdu &25.6 &20.1 &18.5 &17.5 \\
Asturian-Pashto &59.9 &56.5 &56.7 &53.3 \\
Croatian-Afrikaans &0.8 &1.1 &0.9 &0.9 \\
Croatian-Asturian &5.6 &4.2 &3.1 &6.1 \\
Croatian-Urdu &13.2 &16.0 &16.1 &11.8 \\
Croatian-Pashto &50.3 &54.7 &54.5 &48.7 \\
Urdu-Afrikaans &4.4 &4.0 &4.0 &2.1 \\
Urdu-Asturian &24.1 &11.8 &7.9 &9.0 \\
Urdu-Croatian &7.9 &8.5 &8.8 &5.7 \\
Urdu-Pashto &53.4 &59.3 &60.3 &55.2 \\
Pashto-Afrikaans &6.7 &8.3 &7.3 &4.2 \\
Pashto-Asturian &30.7 &17.3 &13.6 &13.2 \\
Pashto-Croatian &16.4 &15.4 &14.9 &12.4 \\
Pashto-Urdu &22.7 &30.5 &31.1 &25.9 \\
\addlinespace
Average &19.5 &19.0 &18.5 &16.6 \\
		\bottomrule
	    \end{tabular}
	\end{adjustbox}
	\caption{\label{app:tab:chrf2} 
	  The percentage of hallucinations (chrF < 20; lower is better) of different pivoting methods for SMaLL100 model on selected language pairs of FLORES-101~\cite{goyal-etal-2022-flores} benchmark.
        }
\end{table}

\section{Results of oscillatory hallucinations based on TNG metric}

\label{app:tng}

\begin{table}[!htb]
	\centering
	\tabcolsep=0.1cm
	\begin{adjustbox}{width=0.48\linewidth}
	\begin{tabular}{lS[table-format=2.1]S[table-format=2.1]S[table-format=2.1]S[table-format=2.1]}
    \toprule
    Language Pairs & \multicolumn{1}{c}{Direct} & \multicolumn{1}{c}{MultiAvg} & \multicolumn{1}{c}{MaxEns} & \multicolumn{1}{c}{EN Pivot}\\
    \midrule
Afrikaans-Asturian &2.2 &2.1 &0.7 &1.3 \\
Afrikaans-Croatian &0.3 &0.3 &0.2 &0.2 \\
Afrikaans-Urdu &1.2 &1.5 &1.1 &0.6 \\
Afrikaans-Pashto &6.7 &6.7 &5.1 &4.7 \\
Asturian-Afrikaans &5.1 &3.0 &3.9 &0.2 \\
Asturian-Croatian &11.9 &1.5 &3.0 &0.4 \\
Asturian-Urdu &15.7 &4.1 &4.3 &0.7 \\
Asturian-Pashto &20.9 &12.8 &11.6 &5.0 \\
Croatian-Afrikaans &0.1 &0.2 &0.2 &0.0 \\
Croatian-Asturian &2.2 &2.0 &1.0 &1.2 \\
Croatian-Urdu &1.5 &1.3 &1.1 &0.5 \\
Croatian-Pashto &7.2 &6.4 &4.9 &4.3 \\
Urdu-Afrikaans &8.2 &3.3 &3.3 &0.9 \\
Urdu-Asturian &15.8 &6.2 &4.0 &2.4 \\
Urdu-Croatian &2.5 &2.6 &2.8 &0.8 \\
Urdu-Pashto &9.3 &11.7 &9.5 &4.2 \\
Pashto-Afrikaans &2.7 &3.6 &3.5 &0.8 \\
Pashto-Asturian &16.0 &6.2 &4.1 &2.0 \\
Pashto-Croatian &3.6 &2.9 &3.0 &1.0 \\
Pashto-Urdu &2.5 &4.8 &4.3 &1.6 \\
\addlinespace 
Average &7.3 &4.5 &4.0 &1.9 \\
\bottomrule
	    \end{tabular}
	\end{adjustbox}
	\caption{\label{app:tab:chrf-tng-avg} 
	   Average percentage~(100\%) of hallucinations (TNG metric; lower is better) of different pivoting methods for M2M100 and SMaLL100 on selected language pairs of FLORES-101.
        }
\end{table}

\begin{table}[!htb]
	\centering
	\tabcolsep=0.1cm
	\begin{adjustbox}{width=0.48\linewidth}
	\begin{tabular}{lS[table-format=2.1]S[table-format=2.1]S[table-format=2.1]S[table-format=2.1]}
    \toprule
    Language Pairs & \multicolumn{1}{c}{Direct} & \multicolumn{1}{c}{MultiAvg} & \multicolumn{1}{c}{MaxEns} & \multicolumn{1}{c}{EN Pivot}\\
    \midrule
Afrikaans-Asturian & 1.8 & 2.6 & 1.1 & 0.5\\
Afrikaans-Croatian &0.5 & 0.49 & 0.2 & 0.3 \\
Afrikaans-Urdu & 1.9 & 2.7 & 1.6 & 0.9\\
Afrikaans-Pashto & 10.4 & 11.6 & 9.1 & 8.0\\
Asturian-Afrikaans & 9.8 & 5.6 & 6.8 & 0.2\\
Asturian-Croatian & 22.6 & 2.5 & 5.2 & 0.4\\
Asturian-Urdu &25.3 & 7.6 & 7.7 & 1.2\\
Asturian-Pashto & 36.8 & 23.6 & 21.3 & 9.2\\
Croatian-Afrikaans & 0.1 & 0.2 & 0.2 & 0.0\\
Croatian-Asturian &2.3 & 2.7 & 1.1 & 0.2\\
Croatian-Urdu &2.2 & 2.1 & 1.7 & 0.7 \\
Croatian-Pashto & 11.8 & 11.1 & 8.5 & 7.5 \\
Urdu-Afrikaans &15.2 & 6.1 & 6.1 & 1.5\\
Urdu-Asturian & 18.2 & 9.0 & 6.8 & 2.1 \\
Urdu-Croatian & 4.1 & 4.8 & 4.9 & 1.2 \\
Urdu-Pashto & 13.3 & 21.0 & 16.8 & 7.4 \\
Pashto-Afrikaans & 4.2 & 4.3 & 4.7 & 1.1 \\
Pashto-Asturian & 12.7 & 6.9 & 5.1 & 1.2\\
Pashto-Croatian & 4.2 & 3.7 & 3.7 & 1.6 \\
Pashto-Urdu & 3.3 & 7.5 & 6.2 & 2.8\\
\addlinespace 
Average & 10.9 & 7.6 &6.7 & 2.8\\
\bottomrule
	    \end{tabular}
	\end{adjustbox}
	\caption{\label{app:tab:chrf-tng-m2m} 
	   The percentage of hallucinations (TNG metric; lower is better) of different pivoting methods for M2M100 model on selected language pairs of FLORES-101~\cite{goyal-etal-2022-flores} benchmark.
        }
\end{table}

\begin{table}[!htb]
	\centering
	\tabcolsep=0.1cm
	\begin{adjustbox}{width=0.48\linewidth}
	\begin{tabular}{lS[table-format=2.1]S[table-format=2.1]S[table-format=2.1]S[table-format=2.1]}
    \toprule
    Language Pairs & \multicolumn{1}{c}{Direct} & \multicolumn{1}{c}{MultiAvg} & \multicolumn{1}{c}{MaxEns} & \multicolumn{1}{c}{EN Pivot}\\
    \midrule
Afrikaans-Asturian & 2.5 & 1.6 & 0.2 & 2.1\\
Afrikaans-Croatian & 0.0 & 0.1 & 0.2 & 0.1 \\
Afrikaans-Urdu & 0.4 & 0.3 & 0.6 & 0.2 \\
Afrikaans-Pashto & 2.9 & 1.8 & 1.2 & 1.3 \\
Asturian-Afrikaans & 0.4 & 0.5 & 0.9 & 0.2\\
Asturian-Croatian & 1.2 & 0.4 & 0.7 & 0.4\\
Asturian-Urdu & 6.1 & 0.7 & 0.9 & 0.2\\
Asturian-Pashto & 5.0 & 2.0 & 1.9 & 0.9 \\
Croatian-Afrikaans & 0.0 & 0.1 & 0.1 & 0.0 \\
Croatian-Asturian & 2.1 & 1.3 & 0.8 & 2.2 \\
Croatian-Urdu & 0.8 & 0.4 & 0.4 & 0.3 \\
Croatian-Pashto & 2.6 & 1.7 & 1.3 & 1.2 \\
Urdu-Afrikaans & 1.1 & 0.6 & 0.4 & 0.3 \\
Urdu-Asturian & 13.4 & 3.3 & 1.2 & 2.6 \\
Urdu-Croatian & 0.8 & 0.4 & 0.7 & 0.4 \\
Urdu-Pashto & 5.3 & 2.3 & 2.1 & 1.0 \\
Pashto-Afrikaans & 1.2 & 2.9 & 2.2 & 0.4 \\
Pashto-Asturian & 19.4 & 5.4 & 3.1 & 2.7\\
Pashto-Croatian & 3.0 & 2.1 & 2.3 & 0.3 \\
Pashto-Urdu & 1.6 & 2.2 & 2.4 & 0.4 \\
\addlinespace 
Average & 3.6 & 1.5 & 1.2 & 0.9 \\
\bottomrule
	    \end{tabular}
	\end{adjustbox}
	\caption{\label{app:tab:chrf-tng-small100} 
	   The percentage of hallucinations (TNG metric; lower is better) of different pivoting methods for SMALL100 model on selected language pairs of FLORES-101~\cite{goyal-etal-2022-flores} benchmark.
        }
\end{table}

\end{appendices}

\end{document}